\RequirePackage{fix-cm}
\documentclass[twocolumn]{svjour3}          
\smartqed  
\usepackage{graphicx}
\usepackage{amsmath,amssymb}
\usepackage{booktabs}
\usepackage{multirow}
\begin{document}

\title{Convolutional Networks with Adaptive Inference Graphs
}


\author{Andreas Veit         \and
        Serge Belongie 
}


\institute{A. Veit \at
              Google Research, New York City \\
              \email{aveit@google.com}           
           \and
           S. Belongie \at
              Department of Computer Science \& Cornell Tech \\ Cornell University, New York \\
              \email{sjb344@cornell.edu}
}

\date{}

\maketitle

\begin{abstract}
Do convolutional networks really need a fixed feed-forward structure? What if, after identifying the high-level concept of an image, a network could move directly to a layer that can distinguish fine-grained differences? Currently, a network would first need to execute sometimes hundreds of intermediate layers that specialize in unrelated aspects. Ideally, the more a network already knows about an image, the better it should be at deciding which layer to compute next. In this work, we propose convolutional networks with adaptive inference graphs (ConvNet-AIG) that adaptively define their network topology conditioned on the input image. Following a high-level structure similar to residual networks (ResNets), ConvNet-AIG decides for each input image on the fly which layers are needed. In experiments on ImageNet we show that ConvNet-AIG learns distinct inference graphs for different categories. Both ConvNet-AIG with 50 and 101 layers outperform their ResNet counterpart, while using $20\%$ and $38\%$ less computations respectively. By grouping parameters into layers for related classes and only executing relevant layers, ConvNet-AIG improves both efficiency and overall classification quality. Lastly, we also study the effect of adaptive inference graphs on the susceptibility towards adversarial examples. We observe that ConvNet-AIG shows a higher robustness than ResNets, complementing other known defense mechanisms.
\end{abstract}

\section{Introduction}
Often, convolutional networks (ConvNets) are already confident about the high-level concept of an image after only a few layers.  This raises the question of what happens in the remainder of the network that often comprises hundreds of layers for many state-of-the-art models. To shed light on this, it is important to note that due to their success, ConvNets are used to classify increasingly large sets of visually diverse categories. Thus, most parameters model high-level features that, in contrast to low-level and many mid-level concepts, cannot be broadly shared across categories. As a result, the networks become larger and slower as the number of categories rises. Moreover, for any given input image the number of computed features focusing on unrelated concepts increases.

What if, after identifying that an image contains a bird, a ConvNet could move directly to a layer that can distinguish different bird species, without executing intermediate layers that specialize in unrelated aspects? Intuitively, the more the network already knows about an image, the better it could be at deciding which layer to compute next. This shares resemblance with decision trees that employ information theoretic approaches to select the most informative features to evaluate. Such a network could decouple inference time from the number of learned concepts. A recent study~\cite{resnetensemble} provides a key insight towards the realization of this scenario. The authors study residual networks (ResNets)~\cite{resnet} and show that almost any individual layer can be removed from a trained ResNet without interfering with other layers. This leads us to the following research question: \emph{Do we really need fixed structures for convolutional networks, or could we assemble network graphs on the fly, conditioned on the input?} 

In this work, we propose ConvNet-AIG, a convolutional network that adaptively defines its inference graph conditioned on the input image. Specifically, ConvNet\nobreakdash-AIG learns a set of convolutional layers and decides for each input image which layers are needed. By learning both general layers useful to all images and expert layers specializing on subsets of categories, it allows to only compute features relevant to the input image. It is worthy to note that ConvNet-AIG does not require special supervision about label hierarchies and relationships to guide layers to specialize.

\begin{figure}[t]
	\begin{center}
		\includegraphics[width=1.0\linewidth]{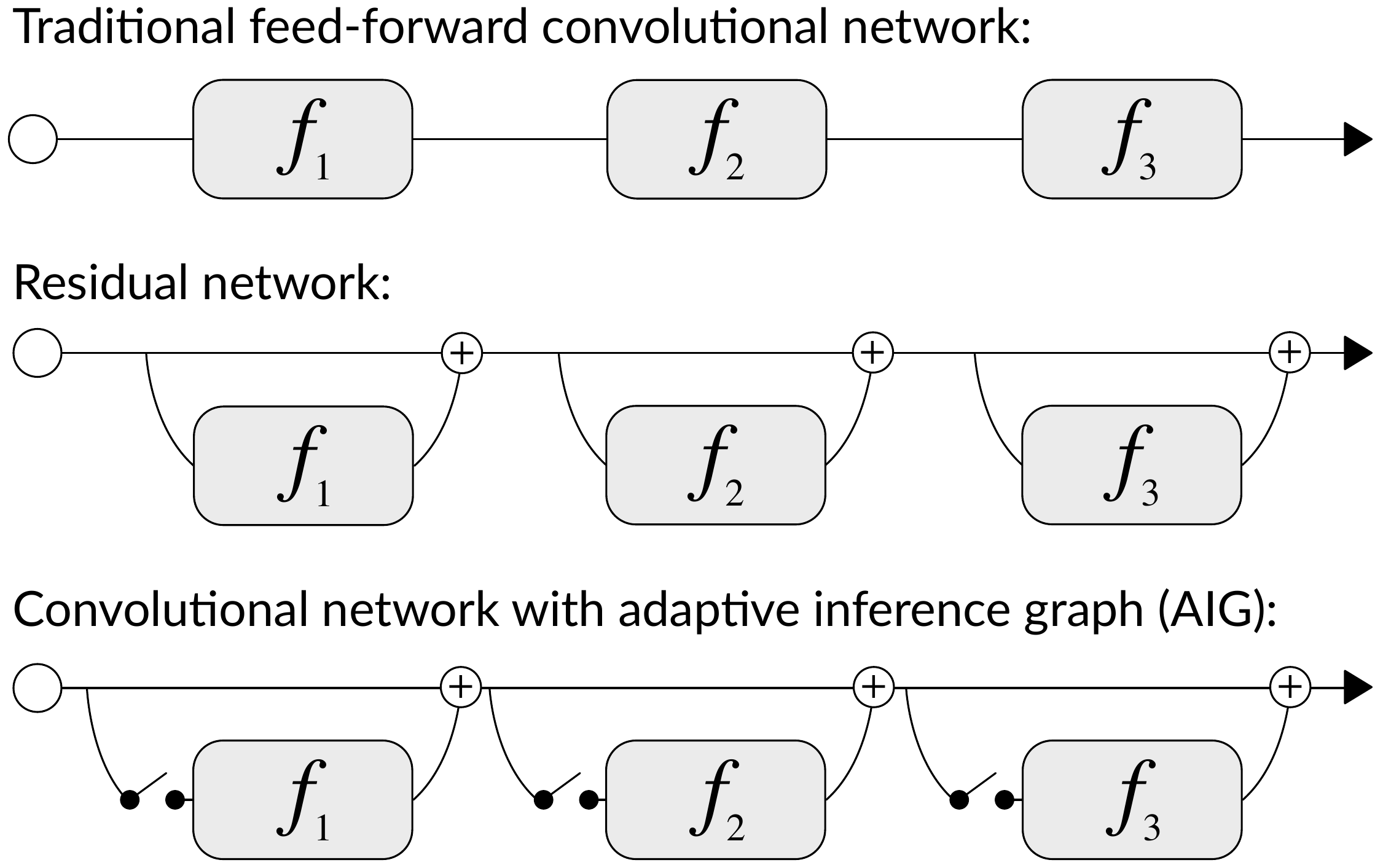}
	\end{center}
	\caption{ConvNet-AIG (bottom) follows a high level structure similar to ResNets (center) by introducing identity skip-connections that bypass each layer. The key difference is that for each layer, a gate determines whether to execute or skip the layer. This enables individual inference graphs conditioned on the input.}
	\label{fig:page1}
\end{figure}

Figure~\ref{fig:page1} gives an overview of our approach. ConvNet-AIG (bottom) follows a structure similar to a ResNet (center). The key difference is that for each residual layer, a gate determines whether the layer is needed for the current input image. The main technical challenge is that the gates need to make discrete decisions, which are difficult to integrate into convolutional networks that we would like to train using gradient descent. To incorporate the discrete decisions, we build upon recent work~\cite{bengio2013estimating,gumbel,concrete} that introduces differentiable approximations for discrete stochastic nodes in neural networks. In particular, we model the gates as discrete random variables over two states: to execute the respective layer or to skip it. Further, we model the gates conditional on the output of the previous layer. This allows to construct inference graphs adaptively based on the input and to train both the convolutional weights and the discrete gates jointly end-to-end.

In experiments on ImageNet~\cite{imagenet}, we demonstrate that ConvNet-AIG effectively learns to generate inference graphs such that for each input only relevant features are computed. In terms of accuracy both ConvNet-AIG~50 and ConvNet-AIG~101 outperform their ResNet counterpart, while at the same time using $20\%$ and $38\%$ less computations. We further show that, without specific supervision, ConvNet-AIG discovers parts of the class hierarchy and learns specialized layers focusing on subsets of categories such as animals and man-made objects. It even learns distinct inference graphs for some mid-level categories such as birds, dogs and reptiles. By grouping parameters for related classes and only executing relevant layers, ConvNet-AIG both improves efficiency and overall classification quality. Lastly, we also study the effect of adaptive inference graphs on susceptibility towards adversarial examples. We show that ConvNet-AIG is consistently more robust than ResNets, independent of adversary strength and that the additional robustness persists even when applying additional defense mechanisms.

\section{Related Work}
Our study is related to work in multiple fields. Several works have focused on \textbf{neural network composition} for visual question answering (VQA) \cite{Andreas2016learning,Andreas2016neural,Johnson2017inferring} and zero-shot learning~\cite{misra2017red}. While these approaches include convolutional networks, they focus on constructing a fixed computational graph up front to solve tasks such as VQA. In contrast, the focus of our work is to construct a convolutional network conditioned on the input image on the fly during execution.

Our approach can be seen as an example of \textbf{adaptive computation} for neural networks. Cascaded classifiers~\cite{viola2004robust} have a long tradition for computer vision by quickly rejecting ``easy'' negatives. Recently, similar approaches have been proposed for neural networks~\cite{li2015convolutional,yang2016exploit}. In an alternative direction,~\cite{bengio2015conditional,shazeer2017outrageously} propose to adjust the amount of computation in fully-connected neural networks. To adapt computation time in convolutional networks,~\cite{Huang2017multi,teerapittayanon2016branchynet} propose architectures that add classification branches to intermediate layers. This allows stopping a computation early once a satisfying level of confidence is reached. Most closely related to our approach is the work on spatially adaptive computation time for residual networks~\cite{figurnov2016spatially}. In that paper, a ResNet adaptively determines after which layer to stop computation. Our work differs from this approach in that we do not perform early stopping, but instead determine which subset of layers to execute. This is key as it allows the grouping of parameters that are relevant for similar categories and thus enables distinct inference graphs for different categories.

Our work is further related to network \textbf{regularization with stochastic noise}. By randomly dropping neurons during training, Dropout~\cite{srivastava2014dropout} offers an effective way to prevent neural networks from over-fitting. Closely related is the work on stochastic depth~\cite{stochasticdepth}, where entire layers of a ResNet are randomly removed during each training iteration. Our work resembles this approach in that it also includes stochastic nodes that decide whether to execute layers. However, in contrast to our work, layer removal in stochastic depth is independent from the input and aims to \emph{increase} redundancy among layers. In our work, we construct the inference graph conditioned on the input image to \emph{reduce} redundancy and allow the network to learn layers specialized on subsets of the data.

Lastly, our work can also be seen as an example of an \textbf{attention mechanism} in that we select specific layers of importance for each input image to assemble the inference graph. This is related to approaches such as highway networks~\cite{highway} and squeeze-and-excitation networks~\cite{hu2017squeeze} where the output of a residual layer is rescaled according to the layer's importance. This allows these approaches to emphasize some layers and pay less attention to others. In contrast to our work, these are soft attention mechanisms and still require the execution of every single layer. Our work is a hard attention mechanism and thus enables decoupling computation time from the number of categories.

\begin{figure}[t]
	\begin{center}
		\includegraphics[width=0.9\linewidth]{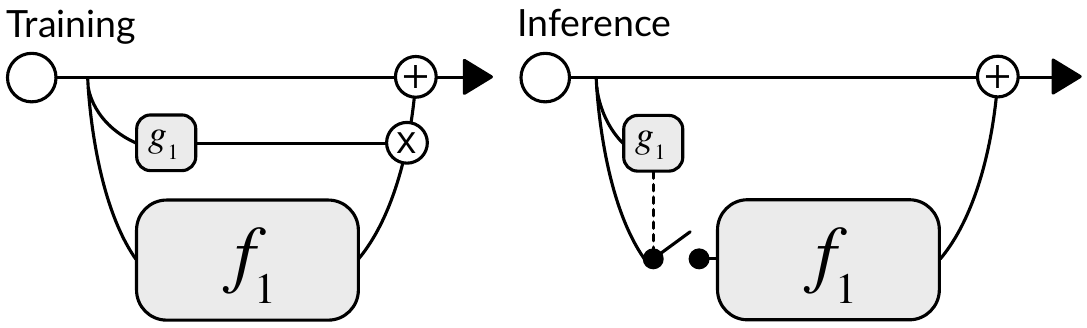}
	\end{center}
	\caption{Left: During training, the output of a gate is multiplied with the output of its respective layer. Right: During inference, a layer does not need to be executed if its gate decides to skip the layer.}
	\label{fig:page2}
\end{figure}

\section{Adaptive Inference Graphs}
Traditional feed-forward ConvNets can be considered as a set of $N$ layers which are sequentially applied to an input image. Figure~\ref{fig:page1} (top) provides an exemplary illustration. Formally, let $f_l(\cdot)$, $l \in \{1, ..., N\}$ denote the function computed by the $l^{th}$ layer. With $\mathbf{x}_0$ as input image and $\mathbf{x}_l$ as output of the $l^{th}$ layer, such a network can be recursively defined as 
\begin{equation}
	\mathbf{x}_{l} = f_l(\mathbf{x}_{l-1})
\end{equation}
ResNets~\cite{resnet}, shown in the center of Figure~\ref{fig:page1}, change this definition by introducing identity skip-connections that bypass each layer, i.e., the input to each layer is also added to its output. This has been shown to greatly ease optimization during training. As gradients can propagate directly through the skip-connection, early layers still receive sufficient learning signal even in very deep networks. A ResNet can be defined as
\begin{equation}
	\mathbf{x}_{l} = \mathbf{x}_{l-1} + f_{l}\left(\mathbf{x}_{l-1}\right)
\end{equation}
In a follow-up study~\cite{resnetensemble} on the effects of the skip-connection, it has been shown that, although all layers are trained jointly, they exhibit a high degree of independence. Further, almost any individual layer can be removed from a trained ResNet without harming performance and interfering with other layers. 

\subsection{Gated Computation}
Inspired by the observations in~\cite{resnetensemble}, we design ConvNet-AIG, a network that can define its topology on the fly. The architecture follows the basic structure of a ResNet with the key difference that instead of executing all layers, the network determines for each input image which subset of layers to execute. In particular, with layers focusing on different subgroups of categories, it can select only those layers necessary for the specific input.  A ConvNet-AIG can be defined as
\begin{equation}
	\begin{split}
		&\mathbf{x}_{l} = \mathbf{x}_{l-1} + g_{l}(\mathbf{x}_{l-1})\cdot f_{l}\left(\mathbf{x}_{l-1}\right) \\
		& \text{where  } g_{l}(\mathbf{x}_{l-1}) \in \{0,1\}
	\end{split}
\label{equ:training}
\end{equation}
where $g_{l}(\mathbf{x}_{l-1})$ is a gate that, conditioned on the input to the layer, decides whether to execute the next layer. The gate chooses between two discrete states: 0 for `off' and 1 for `on', which can be seen as a \emph{hard attention mechanism}.

Since during training gradients are required with respect to all gates' parameters, the computational graphs differ between training and inference time. Figure~\ref{fig:page2} illustrates the key difference between the two settings. During training, each gate and layer is executed and the output of each gate is multiplied with the output of its associated layer according to Equation~\ref{equ:training}. Since $g_{l}(\mathbf{x}_{l-1})$ is binary, activations are only propagated through the network, where gates decide to execute their layers. 
During inference, no gradients are needed. As a consequence, computation can be saved as a layer does not need to be executed if its respective gate decides to skip the layer. The computational graph at inference time can thus be defined as follows
\begin{equation}
\mathbf{x}_{l}=
\begin{cases}
\mathbf{x}_{l-1} & \text{if  } g_{l}(\mathbf{x}_{l-1})=0\\
\mathbf{x}_{l-1} + f_{l}\left(\mathbf{x}_{l-1}\right) & \text{if  } g_{l}(\mathbf{x}_{l-1})=1
\end{cases}
\label{equ:inference}
\end{equation}

\begin{figure*}[t]
	\begin{center}
		\includegraphics[width=0.9\linewidth]{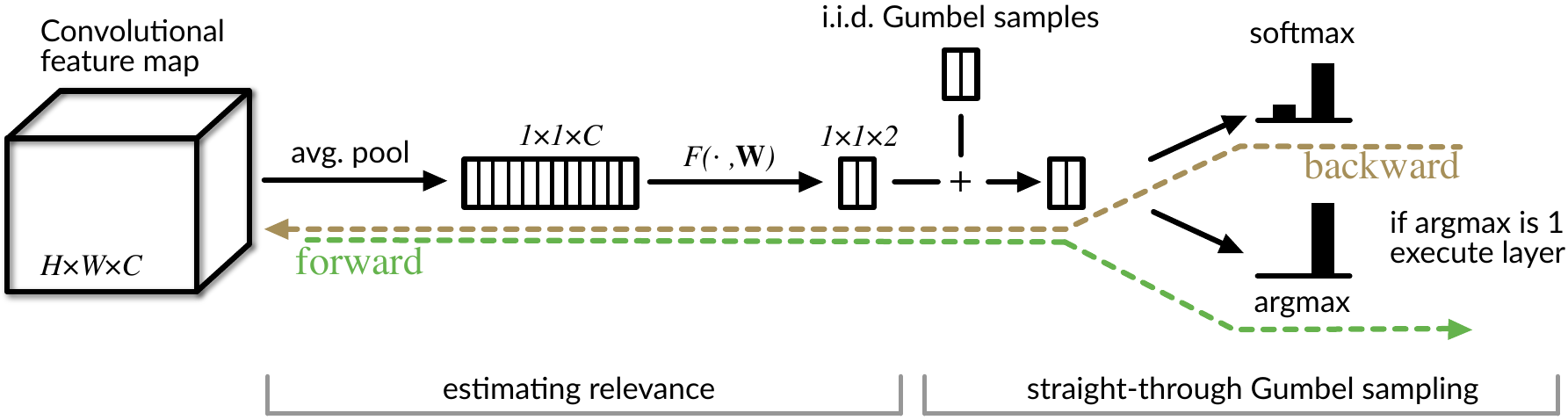}
	\end{center}
	\caption{\textbf{Overview of gating unit.} 
		Each gate comprises two parts. The first part estimates the relevance of the layer to be executed. The second part decides whether to execute the layer given the estimated relevance. In particular, the Gumbel-Max trick and its softmax relaxation are used to allow for the propagation of gradients through the discrete decision.
	}
	\label{fig:details}
\end{figure*}

For the gate to be effective, it needs to address a few key challenges. 
First, to estimate the relevance of its layer, the gate needs to understand its input features. To prevent mode collapse into trivial solutions that are independent of the input features, such as always or never executing a layer, we found it to be of key importance for the gate to be stochastic. We achieve this by adding noise to the estimated relevance. 
Second, the gate needs to make a discrete decision, while still providing gradients for the relevance estimation. We achieve this with the Gumbel-Max trick and its softmax relaxation.
Third, the gate needs to operate with low computational cost. Figure~\ref{fig:details} provides and overview of the two key components of the proposed gate. The first one efficiently estimates the relevance of the respective layer for the current image. The second component makes a discrete decision by sampling using Gumbel-Softmax~\cite{gumbel,concrete}. 

\subsection{Estimating Layer Relevance} The goal of the gate's first component is to estimate the associated layer's relevance given the input features. The input to the gate is the output of the previous layer $\mathbf{x}_{l-1} \in \mathbb{R}^{W\times H \times C}$. Since operating on the full feature map is computationally expensive, we build upon recent studies~\cite{hu2017squeeze,huang2017arbitrary,li2017demystifying} which show that much of the information in convolutional features is captured by the statistics of the different channels and their interdependencies. In particular, we only consider channel-wise means gathered by global average pooling. This compresses the input features into a $1\times 1 \times C$ channel descriptor.
\begin{equation}
	z_c = \frac{1}{H \times W} \sum_{i=1}^{H}\sum_{j=1}^{W} x_{i,j,c}
\end{equation}

To capture the dependencies between channels, we add a simple non-linear function of two fully-connected layers connected with BatchNorm~\cite{batchnorm} and a ReLU~\cite{relu} activation function. The output of this operation is the relevance score for the layer. Specifically, it is a vector $\mathbf{\beta}$ containing two unnormalized scores for the actions of (a) computing and (b) skipping the following layer, respectively.
Generally, a layer is considered relevant for a given input if the score for execution  $\beta_{1}$ is larger than the score for skipping the layer, i.e., $\beta_{0}$. The scores are computed as follows
\begin{equation}
	\mathbf{\beta} = \mathbf{W}_2 \sigma (\mathbf{W}_1 \mathbf{z})
\end{equation}
where $\sigma$ refers to the ReLU, $\mathbf{W}_1 \in\mathbb{R}^{d \times C}$, $\mathbf{W}_2 \in\mathbb{R}^{2 \times d}$ and $d$ is the dimension of the hidden layer. 
The lightweight design of the gating function leads to minimal computational overhead. For a ConvNet-AIG based on Resnet~101 for ImageNet, the gating function adds only a computational overhead of $0.04\%$, but allows to skip $38\%$ of its layers on average.

\subsection{Greedy Gumbel Sampling} The goal of the second component is to make a discrete decision based on the relevance scores. 
For this, we build upon recent work that propose approaches for propagating gradients through stochastic neurons~\cite{bengio2013estimating,kingma2013auto}. In particular, we utilize the Gumbel-Max trick~\cite{gumbelold} and its recent continuous relaxation~\cite{gumbel,concrete}. 

A na\"ive attempt would be to choose the maximum of the two relevance score to decide whether to execute or skip the layer. However, this approach leads to rapid mode collapse as it does not account for the gate's uncertainty. Further, this approach is not differentiable. Ideally, we would like to choose among the options proportional to their relevance scores.  
A standard way to introduce such stochasticity is to add noise to the scores. 

We choose the Gumbel distribution for the noise, because of its key property that is known as the Gumbel-Max trick~\cite{gumbelold}. A random variable $G$ follows a Gumbel distribution if $G=\mu-\log(-\log(U))$, where $\mu$ is a real-valued location parameter and $U$ a sample from the uniform distribution $U \sim \text{Unif}[0,1]$. Then, the Gumbel-Max trick states that if we samples from  $K$ Gumbel distributions with location parameters $\{\mu_{k'}\}_{k'=1}^{K}$, the outcome of the $k^{\text{th}}$ Gumbel is the largest exactly with the softmax probability of its location parameter
\begin{equation}
	P(\text{k is largest}|\{\mu_{k'}\}_{k'=1}^{K}\}) = \frac{e^{\mu_k}}{\sum_{k'=1}^{K}e^{\mu_{k'}}}
	\label{equ:gumbelmaxproperty}
\end{equation}

With this we can parameterize discrete distributions in terms of Gumbel random variables. In particular, let $X$ be a discrete random variable with probabilities $P(X=k)\propto \alpha_k$ and let $\{G_k\}_{k \in \{1,...,K\}}$ be a sequence of i.i.d.~Gumbel random variables with location $\mu=0$. Then, we can sample from the discrete variable $X$ by sampling from the Gumbel random variables
\begin{equation}
	X =  \underset{k \in \{1,...,K\}}{\arg\max} (\log\alpha_k + G_k)
	\label{equ:gumbelmax}
\end{equation}

A drawback of this approach is that the argmax operation is not continuous. To address this, a continuous relaxation of the Gumbel-Max trick has been proposed~\cite{gumbel,concrete}, replacing the argmax with a softmax. Note that a discrete random variable can be expressed as a one-hot vector, where the realization of the variable is the index of the non-zero entry. With this notation, a sample from the Gumbel-Softmax relaxation can be expressed by the vector $\hat{X}$ as follows:
\begin{equation}
	\hat{X}_k = \text{softmax}\left(\left(\log\alpha_k + G_k\right) / \tau \right)
	\label{equ:gumbelsoftmax}
\end{equation}

where $\hat{X}_k$ is the $k^{th}$ element in $\hat{X}$ and $\tau$ is the temperature of the softmax. With $\tau \rightarrow 0$, the softmax function approaches the argmax function and Equation~\ref{equ:gumbelsoftmax} becomes equivalent to the discrete sampler. For $\tau \rightarrow \infty$ it becomes a uniform distribution. Since softmax is differentiable and $G_k$ is independent noise, we can propagate gradients to the probabilities $\alpha_k$. To generate samples with the gating function, we set the log probabilities for the Gumbel-Max trick to the estimated relevance scores, $\log \alpha = \beta$.

One option to employ the Gumbel-softmax estimator is to use the continuous version from Equation~\ref{equ:gumbelsoftmax} during training and obtain discrete samples with Equation~\ref{equ:gumbelmax} during testing.
An alternative is the \emph{straight-through} version~\cite{gumbel} of the Gumbel-softmax estimator. There, during training, for the forward pass we get discrete samples from Equation~\ref{equ:gumbelmax}, but during the backwards pass we compute the gradient of the softmax relaxation in Equation~\ref{equ:gumbelsoftmax}. Note that the estimator is biased due to the mismatch between forward and backward pass. However, we observe that empirically the straight-through estimator performs better and leads to inference graphs that are more category-specific. We illustrate the two different paths during the forward and backward pass in Figure~\ref{fig:details}. 

\subsection{Training Loss}
For the network to learn when to use which layer, we constrain how often each layer is allowed to be used. Specifically, we use soft constraints by introducing an additional loss term that encourages each layer to be executed at a certain target rate. This target rate could be the same for each layer or layer specific. We approximate the execution rates for each layer over each mini-batch and penalize deviations from the target rate. 
Specifically, let $L$ be the set of layers in the network. Each layer has a target rate $t$ which lies within the interval $t_i \in [0, 1]$. Further, with a mini-batch $B$ of training instances $i \in B$ and the output of the gate for the $l^{th}$ layer and $i^{th}$ training instance as $g_{l,i}$, the target rate loss is defined as
\begin{equation}
	\mathcal{L}_{target} = \frac{1}{|L|}\sum_{l\in L} \left(\frac{1}{|B|}\sum_{i\in B}g_{l,i} - t_{l}\right)^{2}
\end{equation}

The target rate loss allows the optimization to reach solutions in which parameters that are relevant only to subsets of related categories are grouped together in separate layers, which minimizes the amount of unnecessary features to be computed.
The target rate provides an easy instrument to adjust computation time. ConvNet-AIG is robust to a wide range of target rates. We study the effect of the target rate on classification accuracy and inference time in the experimental section. With the standard multi-class logistic loss, $\mathcal{L}_{MC}$, the overall training loss is
\begin{equation}
	\mathcal{L}_{AIG} = \mathcal{L}_{MC} + \lambda \mathcal{L}_{target}
\end{equation} 

where $\lambda$ balances the two losses. In our experiments we use $\lambda=2$ We optimize this joint loss with mini-batch stochastic gradient descent.

\subsection{Adaptive Gating During Inference}
Once the network is trained, there are different alternatives for how to perform adaptive gating during inference. The first alternative is to follow the same procedure as used during training and use \emph{stochastic} inference by sampling according to Equation~\ref{equ:gumbelmax}. 
A second alternative is to compute the gates in a \emph{deterministic} fashion. For deterministic inference, we do not sample from the relevance scores by adding Gumbel noise, but directly compute the softmax over them and use a threshold to decide whether to execute the layer
\begin{equation}
g_{l}(\mathbf{x}_{l-1})=
\begin{cases}
0 & \text{if  } \text{softmax}(\beta)_{k'} \leq T\\
1 & \text{if  } \text{softmax}(\beta)_{k'} > T
\end{cases}
\label{equ:deterministic-inference}
\end{equation}
where $k'$ is the element in the relevance score vector $\beta$ that corresponds to executing the layer and $T\in [0, 1]$ is a threshold. With a threshold of $T=0.5$, this performs an argmax over the relevance scores and executes the layer whenever the score for executing is higher that for skipping. 
Thus, varying the threshold, provides a tool that allows to control computation time even after a model has already been trained.
Our empirical evaluation indicates that deterministic inference slightly outperforms the stochastic alternative. Further varying thresholds allows for minor trade-offs between inference time and classification quality.

\section{Experiments}
We perform a series experiments to evaluate the performance of ConvNet-AIG and whether it learns specialized layers and category-specific inference graphs.
We compare the different proposed training and inference modes as well as ablation studies for varying target rates and thresholds.
Lastly, we study its robustness by analyzing the effect of adaptive inference graphs on the susceptibility towards adversarial attacks.

\subsection{Results on CIFAR}
We first perform a set of experiments on CIFAR-10~\cite{cifar10} to validate the proposed gating mechanism and its effectiveness to distribute computation among layers.

\subsubsection{Model configurations and training details} 
We build a ConvNet-AIG based on the original ResNet 110 \cite{resnet}. Besides the added gates, ConvNet-AIG follows the same architecture as ResNet~110. For the gates, we choose a hidden state of size $d=16$. The additional gate per residual block, adds a fixed overhead of $0.01\%$ more floating point operations and $4.8\%$ more parameters compared to the standard ResNet-110. We follow a similar training scheme as~\cite{resnet} with momentum 0.9 and weight decay $5\times 10^{-4}$. All models are trained for 350 epochs with a mini-batch size of 256. We use a step-wise learning rate starting at $0.1$ and decaying by $10^{-1}$ after 150 and 250 epochs. We adopt a standard data-augmentation scheme, where images are padded with 4 pixels on each side, randomly cropped to $32 \times 32$ and with probability 0.5 horizontally flipped.

\subsubsection{Results} 
Table~\ref{tab:cifar} shows test error on CIFAR~10 for ResNet~\cite{resnet}, pre-activation ResNet~\cite{resnet2}, stochastic depth~\cite{stochasticdepth} and their ConvNet-AIG counterpart. The table also shows the number of model parameters and floating point operations (multiply-adds). We compare two variants: For standard ConvNet-AIG, we only execute layers with open gates using the stochastic inference setup. As a second variant, which we indicate by ``~$^{*}$~'', we execute all layers and analogous to Dropout~\cite{srivastava2014dropout} and stochastic depth~\cite{stochasticdepth} the output of each layer is scaled by its expected execution rate. 

From the results, we observe that ConvNet-AIG outperforms its ResNet counterparts clearly, even when using only a subset of the layers. In particular, ConvNet-AIG~110 with a target-rate of $0.7$ uses only $82\%$ of the layers in expectation. Since ResNet~110 might be over-parameterized for CIFAR-10, the regularization induced by dropping layers could be a key factor to performance. We observe that ConvNet-AIG~110$^{*}$ outperforms stochastic depth, implying benefits of adaptive inference graphs beyond regularization. In fact, ConvNet-AIG learns to identify layers of key importance such as downsampling layers and learns to always execute them, although they incur computation cost. We do not observe any downward outliers, i.e.~layers that are dropped every time. 

\begin{table}[t]
\small
	\centering
	\caption{\label{tab:cifar}
		\textbf{Test error on CIFAR~10} in $\%$. ConvNet-AIG~110 clearly outperforms ResNet~110 while only using a subset of $82\%$ of the layers. When executing all layers (ConvNet-AIG~110$^{*}$), it also outperforms stochastic depth. 
	}
		\begin{tabular}{@{}lccc@{}} \toprule
			\textbf{Model} & \textbf{Error} & \textbf{Params $(10^6)$} & \textbf{GFLOPs}\\ \midrule
			ResNet 110~\cite{resnet}  & 6.61 & 1.7 & 0.5\\ 
			Pre-ResNet 110~\cite{resnet2}  & 6.37 & 1.7 & 0.5\\ 
			Stoch. Depth 110  & 5.25 & 1.7 & 0.5\\
			ConvNet-AIG 110  & 5.76 & 1.78 & 0.41\\
			ConvNet-AIG 110$^{*}$  & \textbf{5.14} & 1.78 & 0.5\\ 
			\bottomrule
		\end{tabular}
\end{table}

\subsection{Results on ImageNet}
In experiments on ImageNet~\cite{imagenet}, we study whether the proposed ConvNet-AIG learns to group parameters such that for each image only relevant features are computed. ImageNet is well suited for this study, as it contains a large variety of categories including man-made objects, food, and many different animals.

\subsubsection{Model configurations and training details} 
We build ConvNet-AIGs based on ResNet~50 and ResNet 101~\cite{resnet}. Again, we follow the same architectures as the original ResNets, with the sole exception of the added gates. The size of the hidden state is again $d=16$, adding a fixed overhead of $3.9\%$ more parameters and $0.04\%$ more floating point operations.

\begin{figure}[t]
	\begin{center}
		\includegraphics[width=1.0\linewidth]{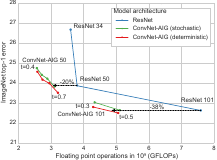}
	\end{center}
	\caption{\textbf{Top-1 accuracy vs. computational cost on ImageNet.} ConvNet-AIG~50 outperforms ResNet~50, while skipping $20\%$ of its layers in expectation. Similarly, ConvNet-AIG~101 outperforms ResNet~101 while requiring $38\%$ less computations. Deterministic inference outperforms stochastic inference, particularly for the larger models with 101 layers.}
	\label{fig:quantitative}
\end{figure}

We compare ConvNet-AIG with different target-rate schedules. First, we evaluate a ConvNet-AIG~50, where all 16 residual layers have the same target rate. As discussed in detail below, in this case some layers are too early in the network to yet effectively distinguish between different categories, and some layers are needed for all inputs. Thus we also evaluate custom target-rate schedules. In particular, for our quantitative results we use a target rate of 1 for all layers up to the second downsampling layer and for ResNet~50 we further set the target rate to 1 for the third downsampling layer and the last layer. 
This slightly improves quantitative performance, and also improves convergence speed.

We follow the standard ResNet training procedure, with mini-batch size of 256, momentum of 0.9 and weight decay of $10^{-4}$. All models are trained for 100 epochs with step-wise learning rate starting at $0.1$ and decaying by $10^{-1}$ every 30 epochs. We use the data-augmentation procedure as in~\cite{resnet} and at test time first rescale images to $256 \times 256$ followed by a $224 \times 224$ center crop. The gates are initialized to open at a rate of $85\%$ at the beginning of training. 

\begin{figure}[t]
	\begin{center}
		\includegraphics[width=1.0\linewidth]{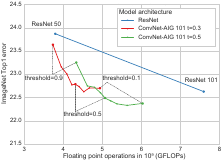}
	\end{center}
	\caption{\textbf{Impact of inference thresholds on top-1 accuracy and computational cost on ImageNet.} Varying thresholds allow for minor trade-offs between inference time and classification quality after a model is already trained. 
For larger adjustments in computation time it is more effective to train a model with a different target-rate.}
	\label{fig:quant-threshold}
\end{figure}

\subsubsection{Quantitative comparison} 
Figure~\ref{fig:quantitative} shows top-1 error on ImageNet and computational cost in terms of GFLOPs for ConvNet-AIG with 50 and 101 layers and the respective ResNets of varying depth. 
We further show the impact of different target rates on performance and efficiency. We compare models with target rates for the layers after the second downsampling layer from $0.4$ to $0.7$ for ConvNet-AIG~50 and $0.3$ to $0.5$ for ConvNet-AIG~101. 
For each variant of ConvNet-AIG we show both the performance of stochastic inference (green) as well as deterministic inference (red). The threshold for each model is set to $0.5$. The impact of varying thresholds is shown for both ConvNet-AIG~101 models in Figure~\ref{fig:quant-threshold}.
Details about the models' complexities and further baselines are presented in Table~\ref{tab:imagenet}. 

\begin{table*}[t]
	\centering
	\caption{\label{tab:imagenet}
		\textbf{Test error on ImageNet} in $\%$ for ConvNet-AIG~50, ConvNet-AIG~101 and the respective ResNets of varying depth. ConvNet-AIGs using the Straight-Through Gumbel training paradigm outperform the standard Gumbel-Softmax, which is indicated with `soft gates'. Further, models with deterministic inference outperform their stochastic counterparts. All numbers shown are based on a threshold of $0.5$. Overall, all ConvNet-AIG variants outperform their ResNet counterpart, while at the same time using only a subset of the layers. This demonstrates that ConvNet-AIG is more efficient and also improves overall classification quality.}
		\begin{tabular}{@{}llcccc@{}} \toprule
			& \textbf{Model} & \textbf{ Top 1 } & \textbf{ Top 5 } & \textbf{ \#Params $(10^6)$ } & \textbf{ FLOPs $(10^9)$}\\ \midrule 
			& ResNet 34~\cite{resnet}  & 26.69 & 8.58 & 21.80 & 3.6\\ 
			& ResNet 50~\cite{resnet}  & 24.7 & 7.8 & 25.56 & 3.8\\
			& ResNet 50 (our)  & 23.87 & 7.12 & 25.56 & 3.8\\
			& ResNet 101~\cite{resnet}  & 23.6 & 7.1 & 44.54 & 7.6\\
			& ResNet 101 (our)  & 22.63 & 6.45 & 44.54 & 7.6\\ \midrule
			& Stochastic Depth ResNet 50  & 27.75 & 9.14 & 25.56 & 3.8\\
			& Stochastic Depth ResNet 101  & 22.80 & 6.44 & 44.54 & 7.6\\ \midrule
			& ConvNet-AIG 50 soft gates [t=0.5]  & 24.42 & 7.30 & 25.56 & 2.95\\
			& ConvNet-AIG 50 soft gates [t=0.7]  & 23.69 & 6.89 & 44.54 & 3.37\\ \midrule
			\multirow{6}*{\textbf{\footnotesize{\rotatebox[origin=c]{90}{stochastic}}}}& ConvNet-AIG 50 [t=0.4]  & 24.75 & 7.61 & 26.56 & 2.56\\
			& ConvNet-AIG 50 [t=0.5]  & 24.42 & 7.42 & 26.56 & 2.71\\
			& ConvNet-AIG 50 [t=0.6]  & 24.22 &7.21 & 26.56 & 2.88\\
			& ConvNet-AIG 50 [t=0.7]  & 23.82 & 7.08 & 26.56 & 3.06\\
			& ConvNet-AIG 101 [t=0.3] & 23.02 & 6.58 & 46.23 & 4.33\\
			& ConvNet-AIG 101 [t=0.5] & 22.63 & 6.26 & 46.23 & 5.11\\ \midrule
			\multirow{6}*{\textbf{\footnotesize{\rotatebox[origin=c]{90}{deterministic}}}}& ConvNet-AIG 50 [t=0.4]  & 24.55 & 7.5 & 26.56 & 2.59\\
			& ConvNet-AIG 50 [t=0.5]  & 24.16 & 7.24 & 26.56 & 2.75\\
			& ConvNet-AIG 50 [t=0.6]  & 23.96 & 7.06 & 26.56 & 2.96\\
			& ConvNet-AIG 50 [t=0.7]  & \textbf{23.5} & \textbf{6.92} & 26.56 & 3.23\\
			& ConvNet-AIG 101 [t=0.3] & 22.78 & 6.54 & 46.23 & 4.31\\
			& ConvNet-AIG 101 [t=0.5] & \textbf{22.48} & \textbf{6.17} & 46.23 & 5.08\\
			\bottomrule
		\end{tabular}
\end{table*}

\begin{figure*}[t]
	\begin{center}
		\includegraphics[width=0.95\linewidth]{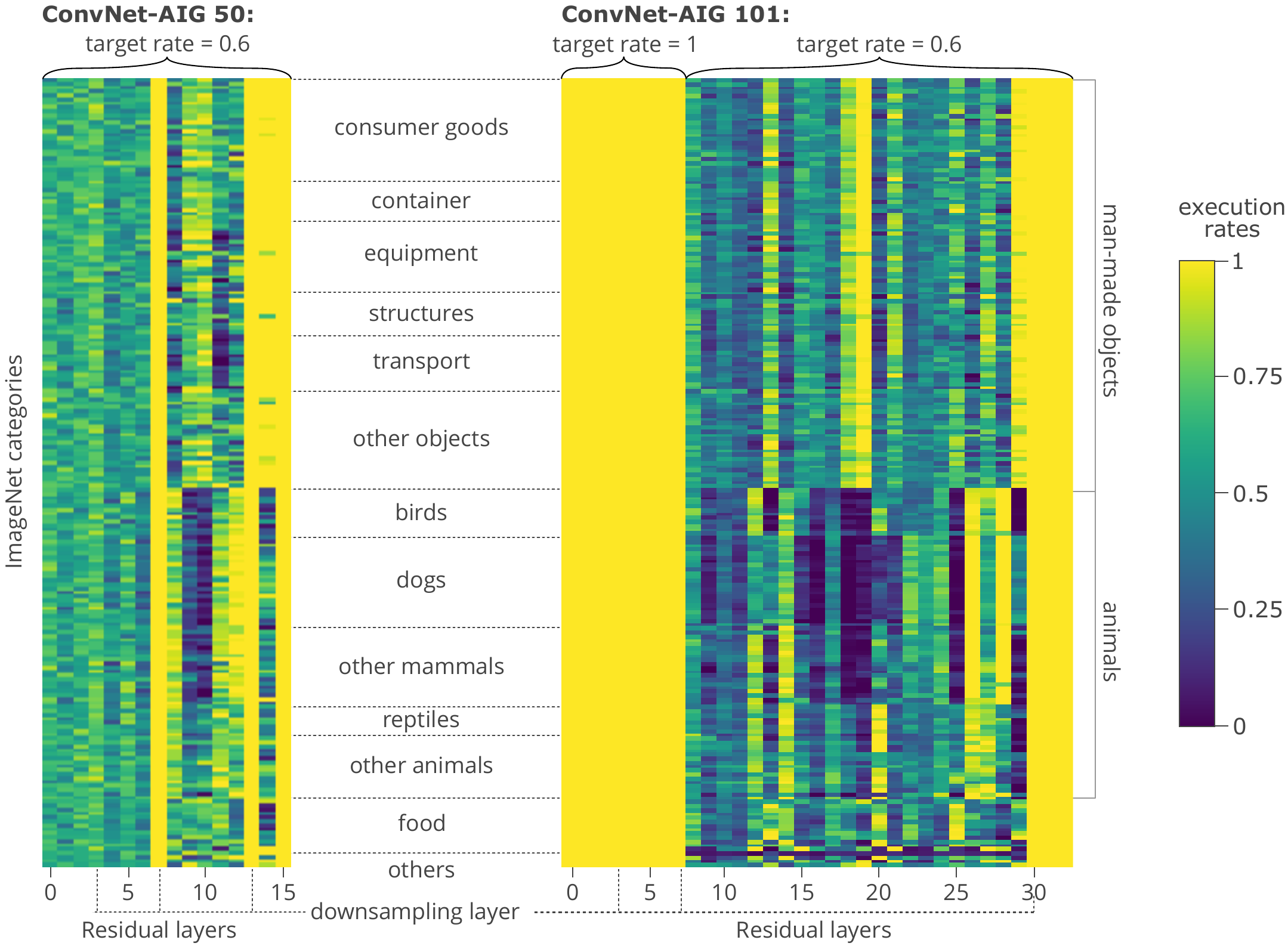}
	\end{center}
	\caption{\textbf{Learned inference graphs on ImageNet.} The histograms show for ConvNet-AIG~50 (left) and ConvNet-AIG~101 (right) how often each residual layer (x-axis) is executed for each of the 1000 classes in ImageNet (y-axis). We observe a clear difference between layers used for man-made objects and for animals and even for some mid-level categories such as birds, mammals and reptiles. Without specific supervision, the network discovers parts of the class hierarchy. Further, downsampling layers and the last layers appear of key importance and are executed for all images. Lastly, the left histogram shows that early layers are mostly agnostic to the different classes. Thus, we set early layers in ConvNet-AIG~101 to be always executed. The remaining layers are sufficient to provide different inference graphs for the various categories.}
	\label{fig:imagenet}
\end{figure*}

From the results we make the following key observations. Both ConvNet-AIG~50 and ConvNet-AIG~101 outperform their ResNet counterpart, while also using only a subset of the layers. In particular, ConvNet-AIG~50 saves about $20\%$ of computation. Similarly, ConvNet-AIG~101 outperforms its respective Resnet while using $38\%$ less computations.
Further, we observe that deterministic inference consistently outperforms stochastic inference. The difference is most noticeable for the large model with 101 layers. This is likely due to the larger proportion of `specialization layers' in the larger model that are focusing on specific subsets of the data, which is highlighted in Figure~\ref{fig:imagenet}.

These results indicate that \emph{convolutional networks do not need a fixed feed-forward structure} and that ConvNet\nobreakdash-AIG is an effective means to enable adaptive inference graphs that are conditioned on the input image.

Figure~\ref{fig:quantitative} also visualizes the effect of the target rate. As expected, decreasing the target rate reduces computation time. Interestingly, penalizing computation first improves accuracy, before lowering the target rate further decreases accuracy. This demonstrates that ConvNet\nobreakdash-AIG both improves efficiency and overall classification quality. Further, it appears often more effective to decrease the target rate compared to reducing layers in standard ResNets.

In Table~\ref{tab:imagenet}, we also compare the two different training regimes, of (a) standard Gumbel-Softmax, where softmax is applied during both forward and backward pass and (b) Straight-Through Gumbel, where argmax is performed during forward and softmax during backward pass. Specifically, we compare performance for ConvNet-AIG 50 with target-rates of $0.5$ and $0.7$. The results show that the straight-through variant consistently outperforms the standard Gumbel-Softmax.
One reason for the observed difference could be that, when using softmax during the forward pass, although scaled down, activations are always propagated through the network, even if a gate decides that a layer is not relevant for a given input image.

Lastly, due to surface resemblance, we also compare our model to stochastic depth~\cite{stochasticdepth}. We observe that for smaller ResNet models stochastic depth does not provide competitive results. Only very large models see benefits from stochastic depth regularization. The paper on stochastic depth~\cite{stochasticdepth} reports that even for the very large ResNet~152 performance remains below a basic ResNet. This highlights the opposite goals of ConvNet\nobreakdash-AIG and stochastic depth. Stochastic depth aims to create redundant features by enforcing each subset of layers to model the whole dataset~\cite{resnetensemble}. ConvNet-AIG aims to separate parameters that are relevant to different subsets of the dataset into different layers.

\subsubsection{Evaluating gating thresholds during inference}

During training, the gates are optimized for a threshold of $0.5$, i.e., argmax over the relevance scores. However, there is some flexibility at test time to vary the thresholds so as to adjust the trade-off between inference time and classification quality. Figure~\ref{fig:quant-threshold} shows top-1 error on ImageNet and computational cost for ConvNet-AIG~101 with target-rates 0.3 and 0.5 for thresholds ranging from $0.1$ to $0.9$.
The results show that varying the thresholds within the range of $0.3$ to $0.7$ allows to slightly adjust inference time of an already trained model without a large decrease in accuracy. However, for larger adjustments to computation time it is more effective to train a model with a different target rate.

\subsubsection{Analysis of learned inference graphs} 
To analyze the learned inference graphs, we study the rates at which different layers are executed for images of different categories. Figure~\ref{fig:imagenet} shows the execution rates of each layer for ConvNet-AIG~50 on the left and ConvNet-AIG~101 on the right. The x-axis indicates the residual layers and the y-axis breaks down the execution rates by the 1000 classes in ImageNet. Further, the figure shows high-level and mid-level categories that contain large numbers of classes. The color in each cell indicates the percentage of validation images from a given category that the respective layer is executed.

\begin{figure*}[t]
	\begin{center}
		\includegraphics[width=1\linewidth]{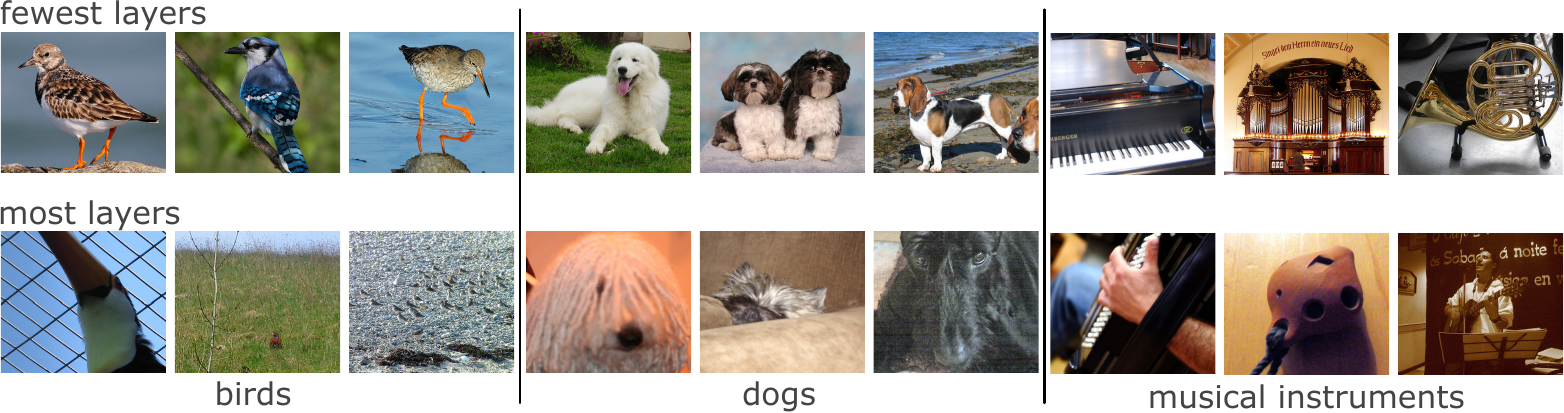}
	\end{center}
	\caption{\textbf{Validation images from ImageNet} that use the fewest layers (top) and the most layers (bottom) within the categories of birds, dogs and musical instruments. The examples illustrate how instance difficulty translates into layer usage.} 
	\label{fig:examples}
\end{figure*}

\begin{figure}[t]
	\begin{center}
		\includegraphics[width=1\linewidth]{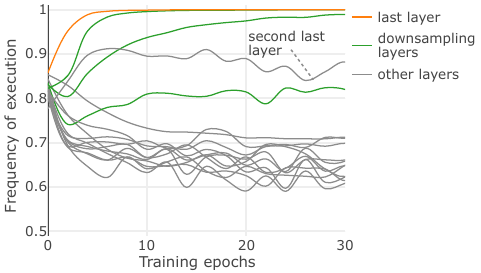}
	\end{center}
	\caption{\textbf{Execution rates per layer over first 30 epochs} of training. Layers are quickly separated into key and less critical layers. Downsampling layers and the last layer increase execution rate, while the remaining layers slowly approach the target rate.}
	\label{fig:overtime}
\end{figure}

\begin{figure}[h]
	\begin{center}
		\includegraphics[width=0.85\linewidth]{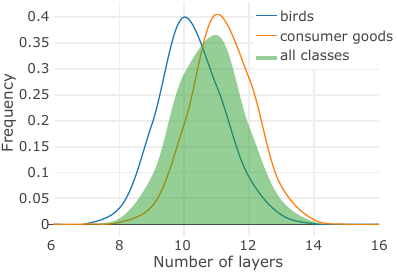}
	\end{center}
	\caption{\textbf{Distribution over the number of executed layers.} For ConvNet-AIG~50 on ImageNet with target rate $0.4$, in average 10.8 out of 16 residual layers are executed. Images of animals tend to use fewer layers than man-made objects.}
	\label{fig:distribution}
\end{figure}

From the figure, we see a clear difference between man-made objects and animals. Moreover, we even observe distinctions between mid-level animal categories such as birds, mammals and reptiles. This reveals that the network discovers part of the label hierarchy and groups parameters accordingly. Generally, we observe similar structures in ConvNet-AIG~50 and ConvNet-AIG~101. However, the grouping of the mid-level categories is more distinct in ConvNet-AIG~101 due to the larger number of layers that can capture high-level features. This result demonstrates that ConvNet-AIG successfully learns layers that focus on specific subsets of categories. It is worthy to note that the training objective does not include an incentive to learn category specific layers. The specialization appears to emerge naturally when the computational budget gets constrained. 

Further, we observe that downsampling layers and the last layers deviate significantly from the target rate and are executed for all images. This demonstrates their key role in the network (as similarly observed in~\cite{resnetensemble}) and shows how ConvNet-AIG learns to effectively trade-off computational cost for accuracy.

Lastly, the figure shows that for ConvNet-AIG~50, inter-class variation is mostly present in the later layers of the network after the second downsampling layer. One reason for this could be that features from early layers are useful for all categories. Further, early layers might not yet capture sufficient semantic information to discriminate between categories. Thus, we set the target rate for the early layers of ConvNet-AIG~101 to 1 so that they are always executed. The remaining layers still provide sufficient flexibility for different inference paths for the various categories.

\begin{figure*}[t]
	\begin{center}
		\includegraphics[width=1\linewidth]{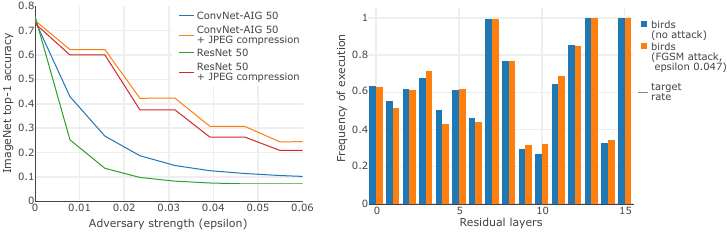}
	\end{center}
	\caption{\textbf{Adversarial attack using Fast Gradient Sign Method. Left:} ConvNet-AIG is consistently more robust than the plain Resnet, independent of adversary strength. The additional robustness persists even when applying additional defense mechanisms. \textbf{Right:} Average execution rates per layer for images of birds before and after the attack. The execution rates remain mostly unaffected by the attack.}
	\label{fig:adversarial}
\end{figure*}

Figure~\ref{fig:overtime} shows a typical trajectory of the execution rates during training for ConvNet-AIG~50 with a target rate $0.6$ for all layers. The layers are initialized to execute a rate of $85\%$ at the start of training. The figure shows the first 30 training epochs and highlights how the layers are quickly separated into key layers and less critical layers. Important layers such as downsampling and the last layers increase their execution rate, while the remaining layers slowly approach the target rate. 

\subsubsection{Variable inference time} 
Due to the adaptive inference graphs, computation time varies across images. Figure~\ref{fig:distribution} shows the distribution over how many of the 16 residual layers in ConvNet-AIG~50 are executed over all ImageNet validation images. On average $10.81$ layers are executed with a standard deviation of $1.11$. The figure also highlights the mid-level categories of birds and consumer goods. It appears that in expectation, images of birds use one layer less than images of consumer goods. From Figure~\ref{fig:imagenet} we further know that the two groups also use different sets of layers. To get a better understanding for what aspects impact inference time, Figure~\ref{fig:examples} shows the validation images that use the fewest and the most layers within the categories of birds, dogs and musical instruments. The examples highlight that easy instances with iconic views require only a few layers. On the other hand, difficult instances that are small or occluded need more computation.

\subsection{Robustness to adversarial attacks}
In a third set of experiments we aim to understand the effect of adaptive inference graphs on the susceptibility towards adversarial attacks. 
Specifically, we are interested in the ConvNet-AIG variant with stochastic inference. 
While exhibiting slightly lower performance compared to the deterministic counterpart, the stochasticity of the inference graph might improve robustness towards adversarial attacks.

We perform a Fast Gradient Sign Attack~\cite{goodfellow2014explaining} on ConvNet\nobreakdash-AIG~50 and ResNet~50, both trained on ImageNet. The results are presented in Figure~\ref{fig:adversarial}. In the graph on the left, the x-axis shows the strength of the adversary measured in the amount each pixel can to be changed. The y-axis shows top-1 accuracy on ImageNet. We observe that ConvNet-AIG is consistently more robust, independent of adversary strength. To investigate whether this additional robustness complements other defenses~\cite{guo2017countering}, we perform JPEG compression on the adversarial examples. We follow~\cite{guo2017countering} and use a JPEG quality setting of 75\%. While both networks greatly benefit from the defense, ConvNet-AIG remains more robust, indicating that the additional robustness can complement other defenses. 
We repeat the experiment for deterministic inference and observe performance very similar to the basic ResNet.

To understand the effect of the attack on the gates and the robustness of stochastic inference, we look at the execution rates before and after the attack. On the right side, Figure~\ref{fig:adversarial} shows the average execution rates per layer over all bird categories for ConvNet-AIG~50 before and after a FGSM attack with epsilon $0.047$. Although the accuracy of the network drops from $74.62\%$ to $11\%$, execution rates remain similar. 
Since the increased robustness only appears during stochastic inference, it seems that the reason for the gates' resilience is the added Gumbel noise which outweighs the noise introduced by the attack.

\section{Conclusion}
In this work, we have shown that convolutional networks do not need fixed feed-forward structures. With ConvNet-AIG, we introduced a ConvNet that adaptively assembles its inference graph on the fly based on the input image. 
Specifically, we presented both a stochastic as well as a deterministic mode to construct the inference graph.
Experiments on ImageNet show that ConvNet-AIG groups parameters for related classes into specialized layers and learns to only execute those layers relevant to the input. This allows decoupling inference time from the number of learned concepts and improves both efficiency as well as overall classification quality. In particular, this translates into $38\%$ less computations for ResNet 101 while achieving the same classification quality. 

This work opens up numerous paths for future work. With respect to network architecture, it would be intriguing to extend this work beyond ResNets to other structures such as densely-connected~\cite{densenet} or inception-based~\cite{googlenet} networks. From a practitioner's point of view, it might be exciting to extend this work into a framework where the set of executed layers is adaptive, but their number is fixed so as to achieve constant inference times. Further, we have seen that the gates are largely unaffected by basic adversarial attacks. For an adversary, it could be interesting to investigate attacks that specifically target the gating functions. 

\begin{acknowledgements}
We would like to thank Ilya Kostrikov, Daniel D. Lee, Kimberly Wilber, Antonio Marcedone, Yiqing Hua and Charles Herrmann for insightful discussions and feedback. 
\end{acknowledgements}

\bibliographystyle{spmpsci}      
\bibliography{egbib}   


\end{document}